\pdfoutput=1
\documentclass[11pt]{article}
\usepackage{acl}
\usepackage{times}
\usepackage{latexsym}
\usepackage[algoruled,ruled,vlined,noend]{algorithm2e}
\usepackage{amsmath}
\usepackage{times}
\usepackage{latexsym}
\usepackage{pgfplots}
\usepackage{graphicx}
\usepackage{enumitem}
\usepackage{tikz}
\usepackage{diagbox}
\usepackage{tkz-tab}
\usepackage{caption}
\usepackage{booktabs}
\usepackage{dsfont}
\usepackage{latexsym}
\usepackage{linguex}
\usepackage{amssymb}
\usepackage{amsmath}
\usepackage{subcaption}
\usepackage{natbib}
\usepackage{blindtext}
\usepackage{url}
\usepackage{color}

\usepackage{url}
\usepackage{hyperref}    
\usepackage{cleveref}
\SetAlFnt{\small}
\SetAlCapFnt{\small}
\SetAlCapNameFnt{\small}
\usepackage{ntheorem}

\usepackage{booktabs, tabularx}
\usepackage{tfrupee}
\usepackage{multicol}
\usepackage{multirow}
\usepackage[font=normalsize,skip=2pt]{caption}
\usepackage{graphicx}
\usepackage{svg}
\usepackage{framed}
\usepackage{MnSymbol,wasysym}
\usepackage{epigraph}
\usepackage[T1]{fontenc}
\usepackage[utf8]{inputenc}
\usepackage{microtype}
\usepackage{dcolumn}
\usepackage{inconsolata}
\usepackage[export]{adjustbox}
\usepackage[colorinlistoftodos,prependcaption,textsize=tiny]{todonotes}
\usepackage[most]{tcolorbox}
\tcbset{on line, 
        boxsep=-0.5pt, arc=1mm, left=0pt,right=0pt,top=0pt,bottom=0pt,
        colframe=white
        }

\title{Memorization Inheritance in Sequence-Level Knowledge\\
Distillation for Neural Machine Translation}

\author{Verna Dankers$^*$ \\
  University of Edinburgh \\
  \texttt{vernadankers@gmail.com} \\\And
  Vikas Raunak$^\ddagger$ \\
  Microsoft \\
  \texttt{viraunak@microsoft.com} \\}

\begin{document}
\newcommand{\appendixshortcut}{Appendix}
\newcommand{\figureshortcut}{Figure}

\maketitle
\hyphenpenalty=500
\begin{abstract}
In this work, we explore how instance-level memorization in 
the \textit{teacher} Neural Machine Translation (NMT) model gets inherited by the \textit{student} model in sequence-level knowledge distillation (SeqKD).
We find that despite not directly seeing the original training data, students memorize more than baseline models (models of the same size, trained on the original data)---3.4\% for exact matches and 57\% for extractive memorization---and show increased hallucination rates.
Further, under this SeqKD setting, we also characterize how students behave on specific training data subgroups, such as subgroups with low quality or specific counterfactual memorization (CM) scores, and find that students exhibit greater denoising on low-quality subgroups.
Finally, we propose a modification to SeqKD named Adaptive-SeqKD, which intervenes in SeqKD to reduce memorization and hallucinations.
Overall, we recommend caution when applying SeqKD: students inherit both their teachers' superior performance \textit{and} their fault modes, thereby requiring active monitoring.
\end{abstract}
\hyphenpenalty=50
\section{Introduction}
\begingroup\def\thefootnote{*}\footnotetext{Work conducted during an internship at Microsoft.}\endgroup
\begingroup\def\thefootnote{$\ddagger$}\footnotetext{Now at Google DeepMind.}\endgroup
Memorization of noisy training data creates unexpected failure modes in \textit{neural machine translation} (NMT) models \citep{raunak2022finding}, thus presenting a reliability risk when deploying them in the real world.
To make NMT models inference-friendly, they are often trained using \textit{sequence-level knowledge distillation} (SeqKD) \citep[e.g.,][]{bapna2022building,costa2022no}. This is a KD variant in which teachers generate synthetic targets for students \citep{hinton2015distilling,kim-rush-2016-sequence}.
\begin{figure}[!h]
\begin{framed}
{\small
\setlength{\Extopsep}{0.5em}
\setlength{\Exlabelsep}{0.75em}
\setlength{\Exlabelwidth}{.75em}
\setlength{\SubExleftmargin}{1.2em}

\noindent \tcbox[colback=pink]{\frownie{}} Extractive Memorization (ExMem) with respect to the initial parallel corpus increases $57.0\%{\pm}15.4$ in students compared to baselines (i.e.\ models memorized to emit the target even if we omit the italicized text):
\ex. \label{ex1}
\a.[$s_C$] Reprezentacja Trynidadu i Tobago \textit{\textcolor{gray}{[w piłce \vfill nożnej]}}
\c.[$t_T$] Trinidad and Tobago national \textcolor{purple}{football} team
\d.[$t_S$] Trinidad and Tobago national \textcolor{purple}{football} team

\noindent \tcbox[colback=pink]{\frownie{}} Students have $31.0\%{\pm}25.7$ more oscillatory hallucinations (in blue) than baselines:
\ex. \label{ex3}
\a.[$s_C$] 1–5 , Stewards are appointed to publish the revelations (\dots)
\b.[$t_T$] 1–5, Diener werden ernannt, um die Offenbarungen zu veröffentlichen (\dots)
\b.[$t_S$] Die Heiligen sind \textcolor{blue}{in der Regel in der Regel in der Regel} (\dots)

\noindent \tcbox[colback=pink]{\frownie{}} Students show secondary ExMem (ExMem with respect to the teacher-generated corpus):
\ex. \label{ex2}
\a.[$s_C$] Electrical industry in Dominican \textit{\textcolor{gray}{[Republic - AmarillasLatinas.net]}}
\b.[$t_T$] Elektrische Industrie in Dominikanische \textcolor{purple}{Republik}
\c.[$t_S$] Elektrische Industrie in Dominikanische \textcolor{purple}{Republik - AmarillasLatinas.net}

\tcbox[colback=lime]{\smiley{}} For low-quality source-target pairs, we observe amplified denoising in students:
\noindent 
\ex. \label{ex4}
\a.[$s_C$] La fiche du Pikauba par la Fromagerie Hamel.
\b.[$t_C$] Pule » Teuerster Käse der Welt aus Eselsmiclh.
\c.[$t_T$] Die Käserei Hamel in Pikauba. \textcolor{gray}{\small (Comet-QE-22${=}0.47$)}
\d.[$t_S$] Die Geschichte des Pikauba durch die Hamel Käserei. \textcolor{gray}{\small (Comet-QE-22${=}0.62$)}
\e.[$t_B$] Die Pikauba-Fassung wird von der Käserei Hamel betrieben. \textcolor{gray}{\small (Comet-QE-22${=}0.47$)}

\vspace{-0.4cm}
}
\end{framed}
\caption{An illustration of our findings. Sources ($s$) are from the corpus ($C$); translations ($t$) are from teachers, students and baselines ($T$, $S$, $B$).}
\vspace{-0.2cm}
\end{figure}
\label{fig:illustration}
SeqKD yields smaller student models whose performance is competitive with the teacher's performance.
Follow-up work has focused primarily on modifying SeqKD objectives to further improve NMT performance \citep[e.g.,][]{wen2023f,zhang-etal-2023-towards-understanding,wang2023better,wang2024don}.
Apart from being used for model compression, SeqKD has proved very beneficial for low-resource and long-tail data \citep{dabre2020combining,currey2020distilling,gumma2023empirical,zhou2024multi,de2024hybrid} but its applications extend beyond that as well, e.g., to continual learning \citep{chuang2020lifelong, zhao2022life}.

Yet, the \textit{understanding} of SeqKD lags behind its \textit{usage}.
Prior work in this direction primarily studies why SeqKD is successful, attributing it to mode reduction of the training data \citep{zhouunderstanding,song2021data}, or suggesting that SeqKD acts as a regularization technique \citep{gordon2019explaining}. In this work, we better try to understand how model behavior gets transmitted from teacher to student, moving beyond only analyzing average-case performance.
In particular, we focus on how the student inherits instance-level memorization.
Recent work on image classification suggests that KD inhibits memorization in the student \citep{lukasiklarger}, but also that membership inference attacks on the student are often successful \citep{jagielski2024students}.
However, the connection between memorization and SeqKD in NLP is new territory; characterizing this is imperative to mitigate memorization-related failures in students.

Our main contributions are as follows: (1) We provide a quantification of \textbf{memorization inheritance} in \S\ref{sec:memorization_inheritance}: we identify that even though SeqKD inhibits memorization from teacher to student, the student memorizes more about the initial parallel corpus and hallucinates more than it would have, had it been trained without SeqKD.
(2) We perform \textbf{subgroup analyses} in \S\ref{sec:subgroup_analysis}, for data subsets with specific characteristics, such as examples of different quality levels or with specific counterfactual memorization scores \citep{feldman2020does}.
We identify subgroups for which the student outperforms both the teacher and baseline models.
(3) To reduce memorization and mitigate accentuated hallucinations we propose \textbf{Adaptive-SeqKD} in \S\ref{sec:Adaptive-SeqKD}: a simple intervention in the SeqKD algorithm wherein we adapt the teacher by finetuning it briefly on \textit{intrinsically} obtained high-quality data to reduce memorization and hallucinations in the student.

Figure~\ref{fig:illustration} demonstrates a subset of these findings with examples from our datasets.
\section{Memorization inheritance}
\label{sec:memorization_inheritance}

\subsection{Experimental setup}
\label{sec:experimental_setup}
In SeqKD, teacher $\theta_T$ is trained on a corpus with source sequences $\mathcal{S}_C$ and targets $\mathcal{T}_C$, and generates translations of $\mathcal{S}_C$ with beam size $k$. The source sequences and the teacher-generated translations ($\mathcal{S}_C$ and $\mathcal{T}_T$) form the training corpus for student $\theta_S$.
We use data from the WMT20 corpus \citep{barrault-etal-2020-findings}, for five language pairs: \textsc{De}-\textsc{En} and \textsc{En}-\textsc{De} (48M), \textsc{Pl}-\textsc{En} and \textsc{En}-\textsc{Pl} (12M), and \textsc{Fr}-\textsc{De} (14M). \appendixshortcut~\ref{ap:data} provides more detail about the WMT corpora, and the validation and test data.

For all languages, a Transformer-large teacher trains for 300k steps on $\mathcal{S}_C$ and $\mathcal{T}_C$, followed by training a Transformer-base student for 100k steps on $\mathcal{S}_C$ and $\mathcal{T}_T$ ($k{=}1$).\footnote{Training duration set to equal the setup of \citet{vaswani2017attention}. We train using \texttt{MarianNMT} \citep{junczys2018marian}. For the full training setup see \appendixshortcut~\ref{ap:data} and our codebase: \url{https://github.com/vernadankers/memseqkd}.} 
We also train Transformer-base directly on $\mathcal{S}_C$ and $\mathcal{T}_C$, to have a baseline ($\theta_{B}$) for what the student model would have memorized when exposed directly to WMT20 data.
In \appendixshortcut~\ref{ap:hyperparams}, we furthermore experiment with varying beam size $k$ and the student's model size.

\paragraph{Model quality metrics}
We first examine the models' quality, to ascertain the SeqKD framework works as intended. 
We report the following reference-based metrics for translations generated with beam size five: \textbf{BLEU} and \textbf{chrF}, 
Translation Error Rate (\textbf{TER}), 
and the \textbf{Comet-20} and \textbf{Comet-22} metrics that use neural methods for translation quality estimation \citep{rei2020comet, rei2022comet}.
We supplement this with the reference-free metrics of \textbf{Comet-QE-20}, \textbf{Comet-QE-22}.

All metrics will be applied to WMT test data, and the reference-free methods are furthermore applied to translations of monolingual data: CommonCrawl data provided by \citet{barrault-etal-2020-findings}, and data from the Pulpo poetry corpus \citep{de2023alberti}, to examine out-of-domain performance.

\begin{figure}
    \centering
    \includegraphics[width=\columnwidth]{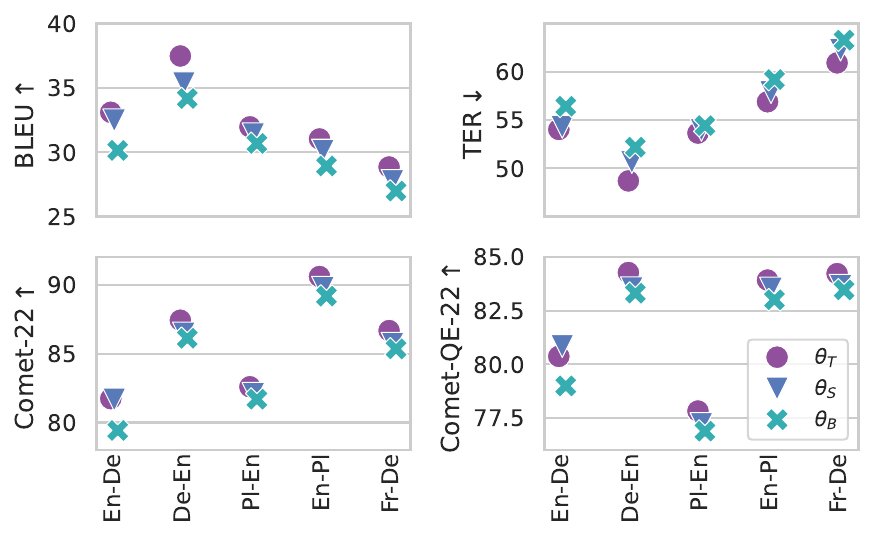}
    \caption{Performance of teacher, student and baseline models for four model quality metrics.}
    \label{fig:model-quality}
    \vspace{-0.3cm}
\end{figure}

\paragraph{Memorization metrics}
We quantify memorization by comparing greedily translated $\mathcal{S}_C$ to $\mathcal{T}_C$ for all models, and to $\mathcal{T}_T$, for $\theta_S$. We measure the \textbf{replication} (exact match) rate, and the \textbf{extractive memorization} \citep[ExMem,][]{raunak2022finding} rate. ExMem finds examples for which models memorized to emit the target after seeing at most $75$\% of the source, e.g., see Example~\ref{ex1}.
The ExMem rate is the percentage of extractively memorized examples out of the replicated examples. 

We also quantify the hallucination rate since hallucinations are often linked to models' memorization capabilities \citep[e.g.,][]{guerreiro2023looking, mckenna2023sources}.
We employ binary metrics because they are high-precision and do not require setting a threshold in the absence of ground truth data \citep{guerreiro2023looking,raunak2021curious}. 
We measure the rates of \textbf{natural hallucinations} (NatHal) and \textbf{oscillatory hallucinations} (OscHal). 
NatHal is the percentage of source sequences that map to a translation that is repeated in the model's translations at least five times.
OscHal is the percentage of translations with bigrams repeated at least 10 times in the target but not the source.\footnote{To improve the memorization/hallucination metrics' precision, some training examples are excluded (see \appendixshortcut~\ref{ap:add_results}, along with implementation details and hyperparameters).} 

\subsection{Results}
\paragraph{General model quality}
\figureshortcut~\ref{fig:model-quality} and \appendixshortcut~\ref{ap:add_results} provide performance differences for $\theta_T$, $\theta_S$ and $\theta_B$.
Overall, $\theta_T$ outperforms $\theta_S$, and $\theta_S$ outperforms $\theta_B$. $\theta_S$ and $\theta_B$ merely differ in the training targets, which demonstrates that our SeqKD pipeline works as intended.
The ordering of models in terms of their quality also holds for CommonCrawl and Pulpo data (see Table~\ref{tab:ap:model_quality}, \appendixshortcut~\ref{ap:add_results}).

\begin{figure}[!t]
    \centering
    \begin{subfigure}[b]{0.49\columnwidth}
        \includegraphics[height=1.53cm, right]{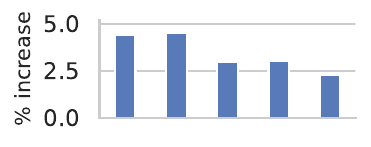}
        \includegraphics[height=3.2cm, right]{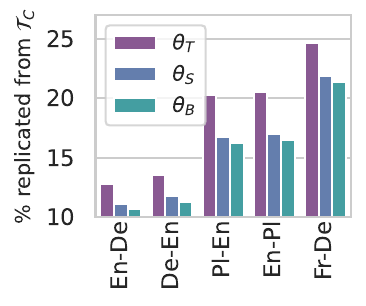}
        \caption{Replication metric}
        \label{fig:mem-metrics-replication}
    \end{subfigure}
    \begin{subfigure}[b]{0.49\columnwidth}
        \includegraphics[height=1.49cm, right]{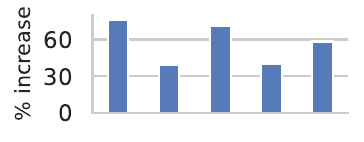}
        \includegraphics[height=3.2cm, right]{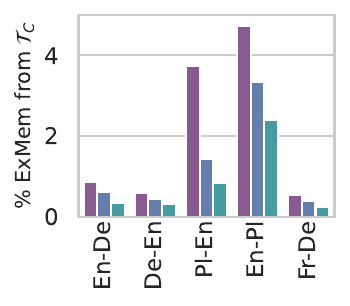}
        \caption{ExMem rate}
        \label{fig:mem-metrics-exmem}
    \end{subfigure}
    \begin{subfigure}[b]{\columnwidth}\centering
        \includegraphics[width=0.9\textwidth]{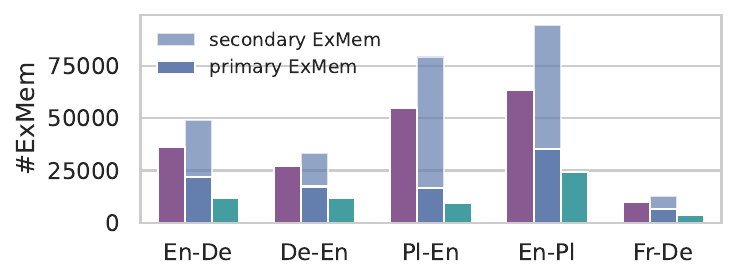}
        \caption{Number of ExMem examples}
        \label{fig:exmem_numbers}
    \end{subfigure}
    \caption{Memorization metrics for $\theta_T$, $\theta_S$ and $\theta_B$ and the percentual increase comparing $\theta_S$ to $\theta_B$.}
    \label{fig:mem-metrics}
    \vspace{-0.2cm}
\end{figure}

\paragraph{SeqKD facilitates memorization}
\figureshortcut~\ref{fig:mem-metrics} summarizes the memorization results.
If we first look at the replication rate, $\theta_S$ replicates less from WMT20 than $\theta_T$ but \textbf{more than $\theta_B$}: students' replication rate with respect to $\mathcal{T}_C$ is 3.4\%($\pm$0.9) higher than for $\theta_B$.
Students also replicate original material from $\theta_T$: the overall student replication rate for $\mathcal{T}_T$ is 35.3\%($\pm2.7$) (see Table~\ref{tab:ap:memorization}, \appendixshortcut~\ref{ap:add_results}).

For the ExMem rate with respect to $\mathcal{T}_C$ (\figureshortcut~\ref{fig:mem-metrics-exmem}), a similar pattern emerges, but with a starker difference between $\theta_S$ and $\theta_B$: students extractively memorize less from $\mathcal{T}_C$ compared to $\theta_T$, but \textbf{more compared to $\theta_B$}, with a mean increase of 57.0\%($\pm$15.4).
This is quite surprising; note that by definition, the rate reported here expresses how many of the replicated examples are extractively memorized. Students only observed 18.4\% (on average) of $\mathcal{T}_C$ through the SeqKD pipeline (the portion that $\theta_T$ replicated) and yet within that smaller pool they still memorized more than $\theta_B$, that was exposed to 100\% of the corpus.
Not only have students extractively memorized WMT20 examples (`primary ExMem'), they also show ExMem with respect to $\theta_T$ (`secondary ExMem', quantified in Table~\ref{tab:ap:memorization}, \appendixshortcut~\ref{ap:add_results}).
\figureshortcut~\ref{fig:exmem_numbers} provides the absolute numbers of ExMem examples, distinguishing primary from secondary ExMem that constitute 41\% and 59\% of all ExMem examples, respectively.
Example~\ref{ex2} demonstrates secondary ExMem: the student has memorized to hallucinate ``AmarillasLatinas.net'' from $\theta_T$'s target when merely shown the source's prefix, but $\theta_T$ has not.\footnote{ExMem is a more widespread issue affecting commercial translation systems too, see Appendix~\ref{ap:commercial}.}

\begin{figure}[!t]
    \centering
    \begin{subfigure}[b]{0.49\columnwidth}
        \includegraphics[height=1.44cm, right]{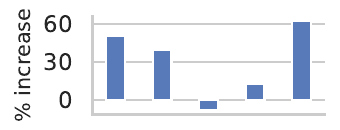}
        \includegraphics[height=3.2cm, right]{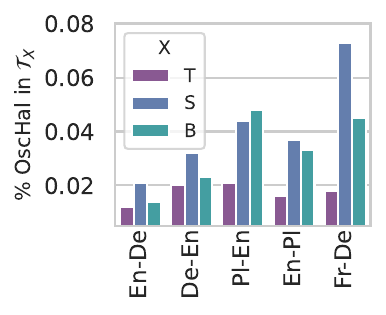}
        \caption{OscHal metric}
        \label{fig:mem-metrics-oschal}
    \end{subfigure}
    \begin{subfigure}[b]{0.48\columnwidth}
        \includegraphics[height=1.54cm, right]{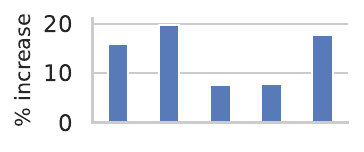}
        \includegraphics[height=3.2cm, right]{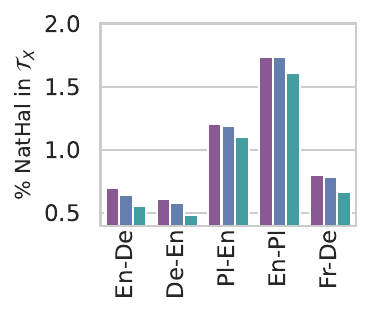}
        \caption{NatHal metric}
        \label{fig:mem-metrics-nathal}
    \end{subfigure}
    \caption{Hallucination metrics for  $\theta_T$, $\theta_S$ and $\theta_B$ and the percentual increase comparing $\theta_S$ to $\theta_B$.}
    \label{fig:hal-metrics}
    \vspace{-0.2cm}
\end{figure}

Why would SeqKD facilitate memorization? We hypothesize that this is due to its denoising function: if noisy data acts as a regularizer during training and $\theta_T$ partly filters that noise through SeqKD, training $\theta_S$ with reduced regularization could lead to increased memorization compared to $\theta_B$. Reduced regularization is traditionally associated with increased memorization in machine learning. In \S\ref{sec:subgroup_analysis}, we demonstrate SeqKD's denoising function.

\paragraph{SeqKD amplifies hallucinations}
It is not just memorization that is amplified through SeqKD; \figureshortcut~\ref {fig:hal-metrics} indicates that \textbf{hallucinations are also amplified}.
For the oscillatory hallucinations, both $\theta_S$ and $\theta_B$ generate more of these than $\theta_T$, and the student hallucinates more than $\theta_B$, on average (an increase of 31.0\%$\pm$25.7, with no increase observed for \textsc{Pl}-\textsc{En}). 
\appendixshortcut~\ref{ap:add_results} presents additional results for monolingual corpora CommonCrawl and Pulpo. Across the board, $\theta_S$ hallucinates most there, too.

For the natural hallucinations, $\theta_S$ and $\theta_B$ hallucinate less than $\theta_T$, but students still hallucinate 13.8\%($\pm$5.0) more than $\theta_B$.

\section{Subgroup analysis}
\label{sec:subgroup_analysis}

We now turn our attention to \textit{data subgroups} to describe SeqKD's impact on samples with specific characteristics.
We compute replication metrics (\textbf{exact match}, \textbf{chrF}), neural quality metrics (\textbf{Comet(-QE)-22}), and textual diversity (\textbf{MSTTR}).
Subgroups contain up to 10k examples, with the exception of the random group.



\begin{figure}[t]\centering
    \begin{subfigure}[b]{0.37\columnwidth}
	\includegraphics[height=3.4cm,left]{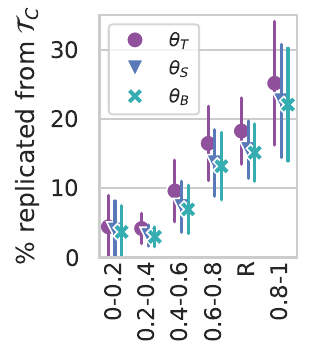}
        \caption{Quality-based}\label{fig:replication_tc_cqe}
    \end{subfigure}
    \begin{subfigure}[b]{0.36\columnwidth}
        \includegraphics[height=3.4cm,left]{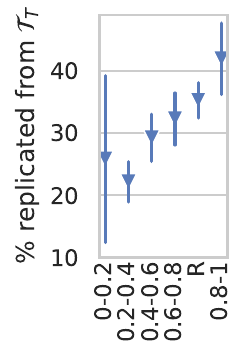}
        \caption{Quality-based}\label{fig:replication_tt_cqe}
    \end{subfigure}
    \begin{subfigure}[b]{0.24\columnwidth}
	\includegraphics[height=3.4cm,right]{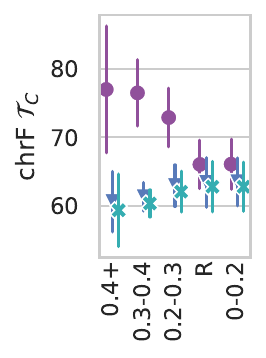}
        \caption{CM-based \small}\label{fig:chrf_cm}
    \end{subfigure}
    \caption{Illustration of how subgroups (indicated in sub-captions) vary in replication. Error bars indicate standard deviations over language pairs.}
    \vspace{-0.4cm}
\end{figure}
\begin{figure}[t]
    \centering
	\includegraphics[width=\columnwidth]{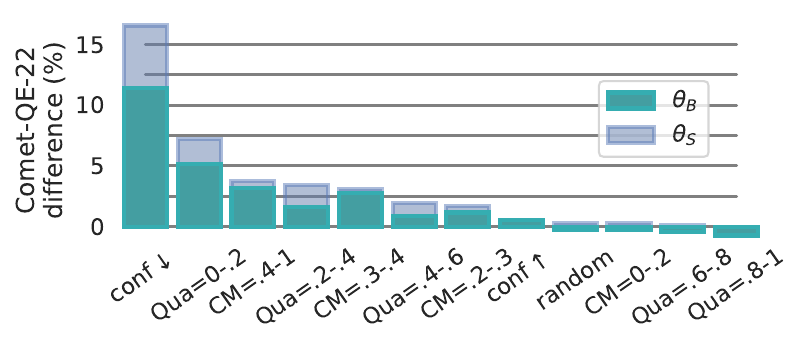}
    \captionof{figure}{Comet-QE-22 increases compared to the teacher per subgroup, averaged over language pairs.}
    \label{fig:subgroups}
    \vspace{-0.1cm}
\end{figure}

\paragraph{Subgroups} We construct 12 subgroups: Firstly, we randomly sample 50k examples from each of the WMT20 corpora.
Secondly, we construct five \textbf{quality-based subgroups} by bucketing WMT20 examples based on the Comet-QE-22 metric, yielding five subgroups (with ranges ${<}$0.2, 0.2-0.4, 0.4-0.6, ${\geq}$0.8).
Thirdly, we create \textbf{counterfactual memorization (CM) subgroups}. Assuming input $x$ and target $y$, and a model with the parameters $\theta^{\text{tr}}$ trained on all training data, and $\theta^{\text{tst}}$ trained on all examples except $(x,y)$, $\texttt{CM}(x, y) {=} p_{\theta^{\text{tr}}}(y|x) - p_{\theta^{\text{tst}}}(y|x)$ \citep{feldman2020does,feldman2020neural}.
We approximate CM scores for all datapoints for \textsc{En}-\textsc{De}, and for 10\% of the datapoints for the remaining language pairs, as detailed in Appendix~\ref{ap:cm}.
We create four subgroups of examples (with ranges ${<}$0.2, 0.2-0.3, 0.3-0.4, ${\geq}$0.4).
Finally, we create two \textbf{confidence-based subgroups}, by taking $\theta_T$'s log-probability averaged over the generated tokens for $\mathcal{T}_T$, selecting the top and bottom 10k examples.

\paragraph{Results}
We refer the reader to \appendixshortcut~\ref{ap:subgroups} for the full results, among which we consider the following patterns to be the most noteworthy:

\noindent \textit{(1) The random subgroup reflects memorization results}: Exact and non-exact match-based metrics for this group (see chrF in \figureshortcut~\ref{fig:random_subgroup}) follow the overall ranking of $\theta_T$>$\theta_S$>$\theta_B$.

\noindent \textit{(2) Quality-based subgroups demonstrate (amplified) denoising}: The lower the quality of the subgroup, the less models replicate from the corpus (\figureshortcut~\ref{fig:replication_tc_cqe},\subref{fig:replication_tt_cqe}). When we look at the Comet-QE-22 quality of the translations the models generate for low-quality subgroups, $\theta_S$ and $\theta_B$ perform better than $\theta_T$, and $\theta_S$ outperforms $\theta_B$ (\figureshortcut~\ref{fig:subgroups}). This effect is thus partially attributable to the capacity gap between large and base models, and partially to KD, and confirms prior work \citep[e.g.,][]{zhouunderstanding} that suggested SeqKD has a denoising function: $\theta_T$ denoised WMT20 for $\theta_S$, and now $\theta_S$ shows \textbf{amplified denoising} on that same data. Example~\ref{ex4} already illustrated this in the introduction.

\noindent \textit{(3) High counterfactual memorization examples are not replicated}: Examples that have high CM for $\theta_T$, do not stand out in terms of replication rates for students or baselines (\figureshortcut~\ref{fig:chrf_cm}), and show some amplified denoising (see \figureshortcut~\ref{fig:subgroups}, although substantially less than low-quality subgroups). 

\noindent \textit{(4) Low-confidence examples are not inherited}: Finally, for examples for which $\theta_T$ generated translations with very low confidence, $\theta_S$ has low replication and strong amplified denoising (\figureshortcut~\ref{fig:subgroups}).

\begin{table}[t]
    \centering\small \setlength{\tabcolsep}{1.5pt}
    \resizebox{\columnwidth}{!}{
    \begin{tabular}{l D{P}{\text{\tiny $\pm$}}{-1} D{P}{\text{\tiny $\pm$}}{-1} D{P}{\text{\tiny $\pm$}}{-1} D{P}{\text{\tiny $\pm$}}{-1}}
    \toprule
    \textbf{Metric} & \multicolumn{1}{c}{$\theta_{T_{hq}}$} & \multicolumn{1}{c}{$\theta_{T_{ra}}$} & \multicolumn{1}{c}{$\theta_{S_{hq}}$} & \multicolumn{1}{c}{$\theta_{S_{ra}}$} \\ \midrule\midrule
    BLEU                 & 0.0 P \text{\tiny $0.5$}    & -1.2 P \text{\tiny $0.8$}  & -0.2 P \text{\tiny $1.7$}   & -1.2 P \text{\tiny $1.6$}   \\
    C-QE-22              & +0.2 P \text{\tiny $0.3$}   & -0.2 P \text{\tiny $0.1$}  & +0.3 P \text{\tiny $0.3$}   & -0.1 P \text{\tiny $0.2$}   \\
    ExMem $\mathcal{T}_C$& -13.1 P \text{\tiny $10.2$} & -7.5 P \text{\tiny $9.3$}  & -23.7 P \text{\tiny $24.0$} & -11.9 P \text{\tiny $29.5$} \\
    OscHal               & -55.1 P \text{\tiny $11.3$} & +7.7 P \text{\tiny $11.2$} & -58.9 P \text{\tiny $7.5$}  & +9.8 P \text{\tiny $15.6$}  \\\bottomrule 
    \end{tabular}}
    \caption{Percentual performance change following Adaptive-SeqKD, compared to $\theta_T$ and $\theta_S$, using high-quality (\textit{hq}) or random (\textit{ra}) data.}
    \label{tab:Adaptive-SeqKD}
    \vspace{-0.1cm}
\end{table}

\section{Adaptive-SeqKD reduces memorization}
\label{sec:Adaptive-SeqKD}
We previously observed increased memorization and hallucinations in $\theta_S$ compared to $\theta_B$, but also amplified denoising on data subsets.
We now investigate whether we can reduce students' failures without harming performance otherwise. 
We apply \textbf{Adaptive-SeqKD}, which briefly finetunes the teacher on high-quality data before producing $\mathcal{T}_T$. Instead of costly external metrics like Comet-QE, we use intrinsic metrics for data quality by selecting 500k sequences where $\theta_T$ nearly memorized the target (\texttt{chrF}>90), $\theta_T$ is confident (normalized translation score > 0.9), and source lengths exceed 5 tokens.
We finetuned $\theta_T$ per language pair for 200 steps and compared it to finetuning on a random 500k sample meeting the length requirement.

Table~\ref{tab:Adaptive-SeqKD} shows that teachers finetuned on the high-quality data (and their students) perform similar in terms of BLEU and Comet-QE, but much better in terms of ExMem and OscHal.
Finetuning on random data reduced ExMem but not hallucinations.
We include additional metrics in \appendixshortcut~\ref{ap:Adaptive-SeqKD}.

Although we applied Adaptive-SeqKD here by finetuning on a training data subset, the technique could be modified for scenarios in which the training data is unknown. For instance, one could simply take a high-quality subset from a new source and finetune models on it prior to running SeqKD.

\section{Related work}

\paragraph{Memorization in NLP and NMT} Memorization as a general topic has seen increased attention in NLP in recent years. A wide range of work examined how many examples large language models memorize and how to extract those memories \citep[e.g.][]{carlini2021extracting, carlini2022quantifying, nasr2023scalable}, and identified how characteristics of datapoints, models and training techniques relate to memorization \citep[e.g.][]{mireshghallah2022empirical, biderman2024emergent, prashanth2024recite, lesci-etal-2024-causal, li2024rome}.
Alternative lines of related work focused on how memorization affects models internally, for instance, for factual memories \citep[e.g.][]{geva2023dissecting} or idiomatic expressions \citep{haviv2023understanding}.

Only a small subset of related work investigated memorization in the NMT context: \citet{raunak2021curious} identified that examples with high CM scores tend to elicit hallucinations from models more easily than other examples. In later work, \citet{raunak2022finding} were the first to discuss the phenomenon of ExMem for NMT systems.
Lastly, \citet{dankers2023memorisation} examined the connection between datapoints' features and CM scores in NMT, and emphasized that memorization can sometimes benefit performance.

\paragraph{Knowledge distillation and memorization}
The connection to memorization and KD had, thus far, only been investigated for the vision domain, where \citet{jagielski2024students} studied membership inference attacks for image classifiers trained through distillation. Although distillation improved the average-case privacy (i.e.\ reduces memorization), the most vulnerable examples barely benefited from KD. \citet{lukasiklarger} studied CM of image classifiers, concluding that distillation inhibits CM. Some of our findings echo these results, namely that compared to $\theta_T$, memorization is indeed dampened in the student, and that high-CM examples do not stand out in terms of replication. 
\section{Conclusion}
\label{sec:conclusion}

SeqKD is popular for effectively training smaller NMT models, but we show that it also introduces issues like worsened ExMem and hallucinations in $\theta_S$ compared to $\theta_B$.
At the same time, the subgroup analyses showed that teachers' CM examples are not necessarily replicated by $\theta_S$, and that students exhibit amplified denoising on low-quality examples.
This highlights a paradox: through SeqKD, students memorize \textit{more} about the corpus than $\theta_B$, yet also outperform $\theta_B$ and $\theta_T$ on data where they \textit{did not} memorize.
Student improvements thus happen both by mimicking $\theta_T$, and by deviating from $\theta_T$.
Future work could suppress memorization during SeqKD, refine Adaptive-SeqKD, and adjust hyperparameters\footnote{Increasing $k$ reduces the student's OscHal rate to below that of $\theta_B$, in particular, with Adaptive-SeqKD yielding further improvements (see \appendixshortcut~\ref{ap:hyperparams}).}
to create more robust SeqKD pipelines.
We advise caution with SeqKD: students may inherit not only the teacher's strengths but also its failures, requiring careful monitoring beyond average-case performance.
\section{Limitations}
We identify the following three limitations with our work:
\begin{itemize}[]
    \item \textbf{Limited experimental setup}: inherent to selecting one (although common) experimental setup of distilling a large model into a smaller model, is that the findings need not necessarily transfer to other settings. We experimented with multiple language pairs, and also varied the beam size and student size in Appendix~\ref{ap:hyperparams} to comment on this limitation, but recognize that there are other settings that would be interesting to study, such as distilling translation models from LLMs. We opted for the more common SeqKD setup, because of its popularity in the years past; most deployed translation systems are not LLM-based (yet). Besides, SeqKD as a technique is still alive and kicking as demonstrated by, for instance, the recently developed OpusDistillery library\footnote{\url{https://github.com/Helsinki-NLP/OpusDistillery}} \citep{opusdistillery}.
    \item \textbf{Niche phenomena}: ExMem and hallucinations are phenomena one would only rarely encounter when interacting with NMT systems \cite{raunak2022finding}. We consider them worthy of investigation, nonetheless, because they represent extreme system failure that goes far beyond a simple mistranslation.
    Particularly for high-resource languages, NMT performance is approaching a ceiling, thanks to the sheer volume of translation examples available. Because NMT test performance is so high, it is important that NMT practitioners go beyond standard evaluation and look into domain-specific phenomena, long-tail phenomena \citep[e.g.][]{raunak-etal-2022-salted}, and robustness failures. ExMem and hallucinations represent such failures.
    \item \textbf{Reliability issues not yet solved}: Adaptive-SeqKD substantially reduced the ExMem and hallucination rates, but did not resolve the issue altogether. We conducted limited experimentation in perfecting Adaptive-SeqKD and recognize this could be further expanded upon in the future. 
\end{itemize}

\section*{Acknowledgments}
We thank the Microsoft Translator team for their valuable input throughout the project, and Matt Post, in particular, for detailed feedback on an earlier version of this article.
VD is supported by the UKRI Centre for Doctoral Training in Natural Language Processing, funded by the UKRI (grant EP/S022481/1) and the University of Edinburgh, School of Informatics and School of Philosophy, Psychology \& Language Sciences, and conducted this work during an internship at Microsoft.
\bibliography{custom}
\bibliographystyle{acl_natbib}
\clearpage
\appendix
\onecolumn
\appendix

\section{Data and experimental setup}
\label{ap:data}
\begin{minipage}{0.52\textwidth}
\paragraph{WMT data}
We download the parallel corpora from the \href{
https://www.statmt.org/wmt20/translation-task.html}{WMT20 website}, using sources listed in Table~\ref{tab:wmt}. For \textsc{En}-\textsc{De} we use the validation/test data from \citet{raunak2022finding}, which is a combination of WMT test data from recent years. For the other language pairs, we use the WMT20 test data, along with the WMT19 test data for validation during training. We honor the licensing terms by using the WMT data for research purposes only and citing the shared task article.
\end{minipage}
\begin{minipage}{0.47\textwidth}
    \small
    \centering
    \begin{tabular}{lrrr}
    \toprule
    \textbf{Source}      & \textsc{En}-\textsc{De} & \textsc{Fr}-\textsc{De} & \textsc{Pl}-\textsc{En}   \\\midrule\midrule
    Europarl    & 1.8M          & 1.8M  & 632k   \\
    ParaCrawl   & 34.4M         & 7.2M  & 6.6M   \\
    Common Crawl& 2.4M          & 622k  & -   \\
    News Commentary& 362k       & 284k  & -   \\
    Wiki Titles & 1.4M          & 942k  & 1.0M   \\
    Tilde Rapid corpus & 1.6M   & -     & 278k   \\
    WikiMatrix  & 6.2M          & 3.4M  & 3.1M   \\
    \bottomrule
    \end{tabular}
    \captionof{table}{Composition of the 3 parallel corpora.}
    \label{tab:wmt}
\end{minipage}

\paragraph{Additional data}
We run monolingual model evaluation using data from additional sources:
\begin{enumerate}[noitemsep,topsep=0pt]
    \item 1M \href{https://www.statmt.org/wmt20/translation-task.html}{\textbf{CommonCrawl}} examples, sampled from the first 100M monolingual CommonCrawl datapoints provided by WMT20. To the best of our knowledge, our use is in line with \href{https://commoncrawl.org/terms-of-use}{CC's terms of use}.
    \item Up to 1M \href{https://huggingface.co/datasets/linhd-postdata/pulpo}{\textbf{Pulpo}} examples, from \citet{de2023alberti}'s multilingual Prolific Unannotated Literary Poetry Corpus containing verses and stanzas. Pulpo contains monolingual sequences for all language pairs, apart from monolingual Polish data. We selected the data because it is expected to be out-of-distribution compared to the WMT20 training corpora. Pulpo contains poems that are copyright-free, or distributed under permissive licenses \citep[][p.3]{de2023alberti}.
\end{enumerate}

\paragraph{Training and evaluation} Models are trained using the \href{https://marian-nmt.github.io/}{\texttt{marian}} toolkit (v1.12.16), using the setup of \citet{raunak2022finding}, that mimics the hyperpameters of \citet{vaswani2017attention}. $\theta_T$ trains for 300k steps; $\theta_S$ and $\theta_B$ train for 100k steps. We use 8 \texttt{Tesla V100-SXM2-32GB} GPUs for model training. 
When evaluating translations using Comet, we use models \texttt{wmt20-comet-da}, \texttt{wmt22-comet-da}, \texttt{wmt20-comet-qe-da} and \texttt{wmt22-cometkiwi-da}, using Comet v1.2.0.
Visit our \href{https://github.com/vernadankers/memseqkd}{git repository} for the training, evaluation and visualization code.

\section{Model quality and memorization}
\label{ap:add_results}
\paragraph{Model quality metrics} Table~\ref{tab:ap:model_quality} provides model quality metrics for all language pairs.
Generally, $\theta_T$>$\theta_S$>$\theta_B$ (except for TER, that should be minimized instead of maximized), and where the results differ we italicized the numbers; this mostly happens for the monolingual CommonCrawl data. Yet, across the board, the pattern is clear for both WMT20 and monolingual data.
The Pulpo scores are noticeably lower for both Comet-QE metrics. This could indicate that the poetry data is harder to translate, but we cannot reliably make a direct comparison across domains. Most importantly, the system ranking holds for this OOD data, too.

\vspace{.3cm}
\noindent\begin{minipage}{\textwidth}
    \begin{center}
    {\centering\small\setlength{\tabcolsep}{2.3pt}
    \begin{tabular}{llccccccccccc}
    \toprule
    \textbf{Lang.} & \textbf{$\theta$} & \multicolumn{7}{c}{\textbf{WMT20}} & \multicolumn{2}{c}{\textbf{CommonCrawl}} & \multicolumn{2}{c}{\textbf{Pulpo}} \\
                     &   & \texttt{BLEU} & \texttt{chrF} & \texttt{TER} & \texttt{Com-20} & \texttt{Com-22} & \texttt{Com-QE-20} & \texttt{Com-QE-22} & \texttt{Com-QE-20} & \texttt{Com-QE-22} & \texttt{Com-QE-20} & \texttt{Com-QE-22} \\\midrule\midrule 
    \multirow{3}{*}{\textsc{En}-\textsc{De}}  & $\theta_T$   & 33.11 & 61.46 & 54.04 & \textit{41.77} & 81.73 & 31.52 & \textit{80.37} & 35.39          & \textit{73.11} & 11.56 & 65.83 \\
                                              & $\theta_{S}$ & 32.49 & 61.01 & 54.31 & \textit{42.00} & 81.66 & 31.11 & \textit{80.85} & 34.93          & \textit{74.00} & 10.86 & 65.37 \\
                                              & $\theta_{B}$ & 30.15 & 59.41 & 56.44 & \textit{34.35} & 79.42 & 28.08 & \textit{78.99} & 34.77          & \textit{72.97} & 10.58 & 63.82 \\\midrule
    \multirow{3}{*}{\textsc{De}-\textsc{En}}  & $\theta_T$   & 37.49 & 64.95 & 48.70 & 68.97          & 87.42 & 49.40 & 84.28          & 38.19          & 77.01          & 9.46  & 64.72 \\
                                              & $\theta_{S}$ & 35.36 & 63.65 & 50.65 & 65.87          & 86.46 & 46.27 & 83.55          & 37.02          & 76.77          & 8.37  & 63.29 \\
                                              & $\theta_{B}$ & 34.18 & 63.11 & 52.20 & 64.68          & 86.11 & 45.39 & 83.32          & 37.02          & 76.55          & 8.36  & 62.54 \\ \midrule
    \multirow{3}{*}{\textsc{Pl}-\textsc{En}}  & $\theta_T$   & 31.98 & 59.14 & 53.66 & 54.20          & 82.59 & 34.70 & 77.82          & \textit{34.35} & \textit{74.58} & n/a   & n/a \\
                                              & $\theta_{S}$ & 31.41 & 58.95 & 53.99 & 51.83          & 82.08 & 32.54 & 77.21          & \textit{34.08} & \textit{74.72} & n/a   & n/a \\
                                              & $\theta_{B}$ & 30.69 & 58.58 & 54.44 & 50.55          & 81.69 & 32.23 & 76.88          & \textit{34.33} & \textit{74.44} & n/a   & n/a \\ \midrule
    \multirow{3}{*}{\textsc{En}-\textsc{Pl}}  & $\theta_T$   & 31.04 & 59.86 & 56.91 & 92.07          & 90.59 & 74.56 & 83.93          & \textit{33.71} & \textit{70.20} & 12.24 & 61.58 \\
                                              & $\theta_{S}$ & 30.13 & 58.98 & 57.91 & 89.09          & 89.77 & 71.99 & 83.53          & \textit{34.18} & \textit{71.08} & 11.46 & 61.16 \\
                                              & $\theta_{B}$ & 28.94 & 58.27 & 59.21 & 86.52          & 89.17 & 70.41 & 83.01          & \textit{34.18} & \textit{70.09} & 10.94 & 59.55 \\ \midrule
    \multirow{3}{*}{\textsc{Fr}-\textsc{De}}  & $\theta_T$   & 28.86 & 60.70 & 60.94 & 60.16          & 86.67 & 48.24 & 84.23          & 33.86          & 72.24          & 9.89  & 58.44 \\
                                              & $\theta_{S}$ & 27.77 & 60.22 & 62.31 & 57.08          & 85.71 & 46.60 & 83.66          & 33.21          & 72.04          & 9.42  & 58.38 \\
                                              & $\theta_{B}$ & 26.99 & 59.69 & 63.30 & 55.95          & 85.34 & 46.06 & 83.47          & 32.90          & 71.58          & 9.03  & 57.30 \\
    \bottomrule
    \end{tabular}}
    \end{center}
    \captionof{table}{Model performance for the five language pairs, using the experimental setup described in \S\ref{sec:experimental_setup}.
    }
    \label{tab:ap:model_quality}
\end{minipage}

\clearpage
\paragraph{Memorization-related metrics} Table~\ref{tab:ap:memorization} provides memorization metrics for all language pairs.
For replication, ExMem and NatHal rates, it holds that $\theta_T$>$\theta_S$>$\theta_B$, whereas for oscillatory hallucinations, $\theta_S$ typically hallucinates most, although there are exceptions (highlighted in italics).

\vspace{.1cm}
\noindent\begin{minipage}{\textwidth}
    \begin{center}
    {\centering\small\setlength{\tabcolsep}{4.5pt}
    \begin{tabular}{llcccccccc}
    \toprule
    \textbf{Lang.} & \textbf{$\theta$} & \multicolumn{6}{c}{\textbf{WMT20}} & \textbf{CommonCrawl} & \textbf{Pulpo} \\
    && \texttt{Replic.} $\mathcal{T}_C$ & \texttt{Replic.} $\mathcal{T}_T$ & \texttt{ExMem} $\mathcal{T}_C$ (\#) & \texttt{ExMem} $\mathcal{T}_T$ (\#) & \texttt{NatHal} & \texttt{OscHal} 
 & \texttt{OscHal} & \texttt{OscHal}\\ \midrule\midrule 
    \multirow{3}{*}{\textsc{En}-\textsc{De}}  & $\theta_T$          & 12.75 &  n/a    &  0.875 (36k) &  n/a         & 0.699 & 0.012         & 0.020          & \textit{0.008}  \\
                                              & $\theta_{S}$        & 11.12 &  32.75  &  0.627 (22k) &  0.397 (49k) & 0.648 & 0.021         & 0.029          & \textit{0.005}  \\
                                              & $\theta_{B}$        & 10.65 &  n/a    &  0.356 (12k) &  n/a         & 0.559 & 0.014         & 0.021          & \textit{0.009}  \\\midrule
    \multirow{3}{*}{\textsc{De}-\textsc{En}}  & $\theta_T$          & 13.53 &  n/a    &  0.590 (27k) &  n/a         & 0.614 & 0.020         & 0.024          & \textit{0.062}  \\
                                              & $\theta_{S}$        & 11.80 &  35.17  &  0.446 (17k) &  0.246 (34k) & 0.583 & 0.032         & 0.037          & \textit{0.048}  \\
                                              & $\theta_{B}$        & 11.29 &  n/a    &  0.320 (12k) &  n/a         & 0.487 & 0.023         & 0.026          & \textit{0.024}  \\\midrule
    \multirow{3}{*}{\textsc{Pl}-\textsc{En}}  & $\theta_T$          & 20.31 &  n/a    &  3.744 (55k) &  n/a         & 1.203 & \textit{0.021}& \textit{0.106} & n/a    \\
                                              & $\theta_{S}$        & 16.73 &  34.65  &  1.433 (17k) &  3.143 (79k) & 1.192 & \textit{0.044}& \textit{0.135} & n/a    \\
                                              & $\theta_{B}$        & 16.25 &  n/a    &  0.836 (9k)  &  n/a         & 1.107 & \textit{0.048}& \textit{0.136} & n/a    \\\midrule
    \multirow{3}{*}{\textsc{En}-\textsc{Pl}}  & $\theta_T$          & 20.56 &  n/a    &  4.736 (63k) &  n/a         & 1.737 & 0.016         & \textit{0.078} & 0.017  \\
                                              & $\theta_{S}$        & 17.02 &  33.64  &  3.347 (35k) &  4.126 (94k) & 1.737 & 0.037         & \textit{0.100} & 0.029  \\
                                              & $\theta_{B}$        & 16.52 &  n/a    &  2.394 (24k) &  n/a         & 1.610 & 0.033         & \textit{0.126} & 0.027  \\\midrule
    \multirow{3}{*}{\textsc{Fr}-\textsc{De}}  & $\theta_T$          & 24.62 &  n/a    &  0.546 (10k) &  n/a         & 0.799 & 0.018         & 0.025          & 0.190  \\
                                              & $\theta_{S}$        & 21.88 &  40.39  &  0.407 (6k)  &  0.330 (13k) & 0.784 & 0.073         & 0.082          & 0.223  \\
                                              & $\theta_{B}$        & 21.39 &  n/a    &  0.257 (4k)  &  n/a         & 0.666 & 0.045         & 0.046          & 0.081  \\
    \bottomrule
    \end{tabular}}\end{center}
    \captionof{table}{Memorization metrics for the five language pairs, using the experimental setup described in \S\ref{sec:experimental_setup}.}
    \label{tab:ap:memorization}
\end{minipage}
\vspace{.0cm}

\paragraph{Additional information on ExMem and hallucination metrics}
To improve the precision of the ExMem and hallucination metrics, some groups of examples are excluded from the computation:
\begin{itemize}[nosep]
\item \textbf{ExMem}: in some cases, it is justified that the model emits the target after having processed only 75\% of the source, e.g., if the target paraphrases the source. To improve the precision, we thus exclude the following examples when computing ExMem: a) examples with a source shorter than 4 words, b) examples with incorrect source or target languages, c) examples for which the length ratios between source and target are over 1.3, d) examples for which the source equals the target.
\item \textbf{OscHal}: examples with source sequences of at least 50 white-space tokenized tokens are excluded, because it becomes more likely that the repeated bigrams might be accurate for longer sequences. This only concerns a small portion of the training data, e.g., only 3.4\% for \textsc{En-De}. We count a sequence as a hallucination if the most frequent bigram appears more than 10 times in the translation, and at least 4 times more often than in the source. We experimentally verified that when reducing that maximum count of 10, the OscHal rate increases, but the model ranking remains the same.
\item \textbf{NatHal}: for the natural hallucinations, we exclude examples for which the Comet-QE-22 score for the source and the generated translation is above 0.85. This is to exclude cases where natural hallucinations are detected simply because there are source sequences that are each other's paraphrase, for instance if both ``Thank you for your visit at our website.'' and ``Thanks for visiting our website.'' map to ``Vielen Dank für Ihren Besuch auf unserer Website.''.
\end{itemize}

\vspace{-0.2cm}

\section{Varying SeqKD hyperparameters}
\label{ap:hyperparams}

Figure~\ref{fig:beam_size} demonstrates how model quality and memorization metrics change when we increase the beam size used to decode $\mathcal{T}_T$ (for both \textsc{Fr}-\textsc{De} and \textsc{En}-\textsc{De}), or change the student's model size (for \textsc{En}-\textsc{De}). $\theta_{S_{L}}$ and $\theta_{S_{S}}$ have hidden dimensionalities of 1024 and 256, respectively (with corresponding feedforward sizes of 4096 and 1024).
When increasing the beam size: i) the quality of the students' translations decrease (for \textsc{En}-\textsc{De}) or remain mostly stable (for \textsc{Fr}-\textsc{De}), ii) the replication rates slightly increase, and iii) ExMem decreases, although not consistently: for \textsc{Fr}-\textsc{De}, the students' ExMem rates all exceed the baseline, whereas for \textsc{De}-\textsc{En} and $k\in\{2,5\}$ the students are slightly below $\theta_B$. This is still concerning, considering that students were exposed to 18.4\% (on average) of the original corpus, whereas the baseline has seen 100\%, so even similar ExMem rates between $\theta_S$ and $\theta_B$ suggest KD facilitates memorization.
The NatHal rate slightly reduces but still far exceeds the baseline. The only real improvement larger beam sizes appear to make are reducing OscHal, both on WMT20 data (OscHal), and on external data
\noindent\begin{minipage}{\textwidth}
    \centering
    \includegraphics[width=.9\textwidth]{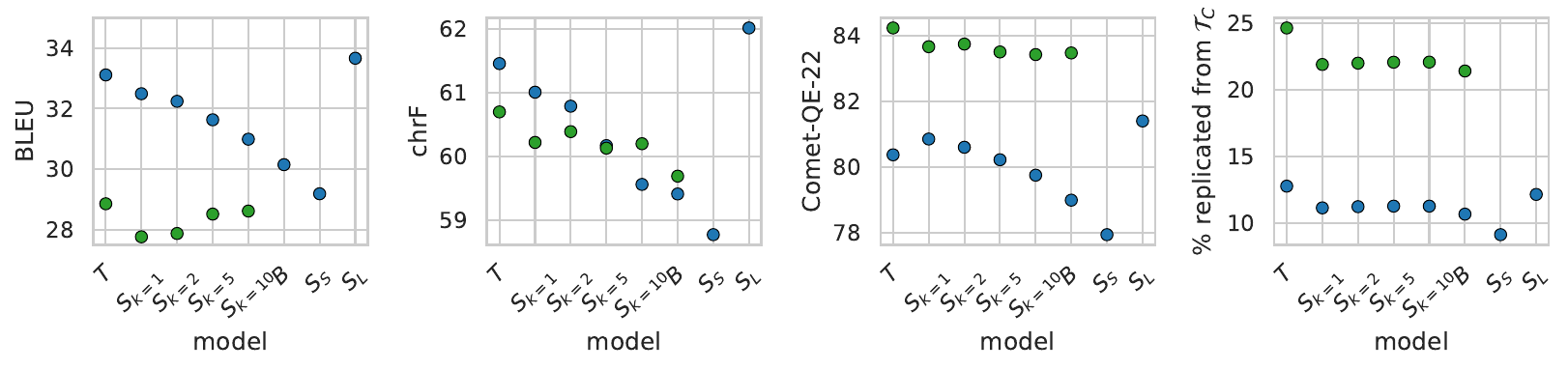}
    \includegraphics[width=.9\textwidth]{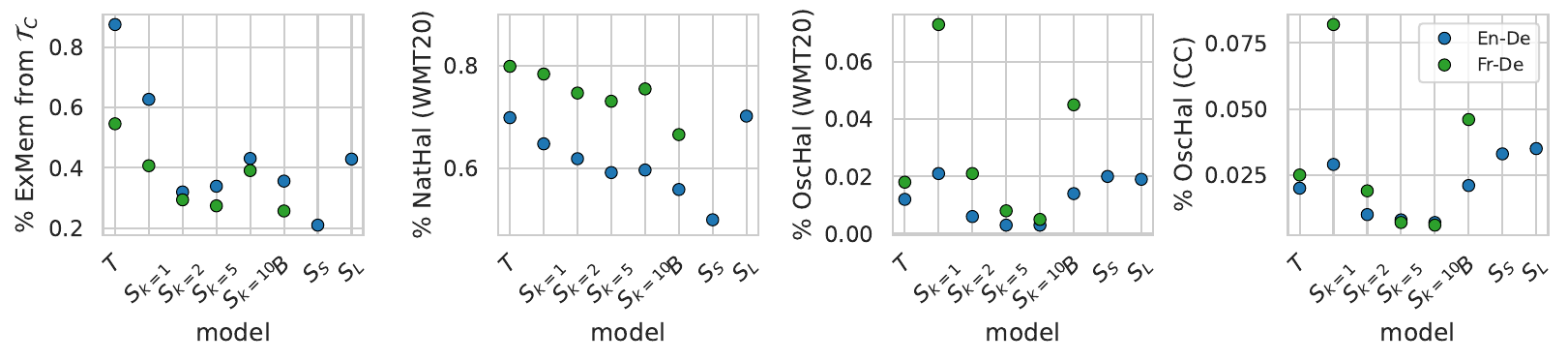}
    \captionof{figure}{Illustration of how model quality and memorization metrics change as a result of changing the SeqKD beam size for \textsc{En}-\textsc{De}.}
    \label{fig:beam_size}
\vspace{0.2cm}
\end{minipage}
(OscHal CommonCrawl). With slight performance reduction in \textsc{En-De} and no performance reduction in \textsc{Fr-De} this yields a simple KD lesson: even if beam search is more computationally expensive (particularly when applied to the \textit{entire} training corpus): do not use greedy search in KD.

Reducing the student's model size reduces the model's translation quality, and reduces replication, ExMem and NatHal, but still yields similar OscHal rates. The larger model produces better translations than its teacher and has lower ExMem, but also increased hallucinations.

\section{Adaptive-SeqKD}
\label{ap:Adaptive-SeqKD}

Figure~\ref{fig:adaptivekd_perlanguage} displays how performance changes per language pair due to Adaptive-SeqKD. Changes to quality metrics like BLEU and Comet(-QE)-22 are relatively minor, but the ExMem and hallucination rates are strongly affected. Finetuning with the high-quality data (\textit{hq} in the graphs) decreases ExMem (for four of five language pairs) and hallucinations (all language pairs); finetuning with random data (\textit{ra} in the graphs) is somewhat effective in reducing ExMem, but mostly fails to effectively reduce the hallucination rates.

In the previous section, we observed that increasing the SeqKD beam size is beneficial for the reduction of the hallucination rate. We thus reran finetuning with high-quality data for $S_{k=5}$ for \textsc{En}-\textsc{De} to examine whether finetuning also helps for an increased beam size. Upon doing so, the OscHal rate reduced with 28\% for the WMT20 data and 33\% for the CommonCrawl data, and the NatHat rate reduced with 10\% for the WMT20 data, suggesting the wider applicability of Adaptive-SeqKD.

\begin{center}
\noindent \begin{minipage}{\textwidth}\begin{center}
    \includegraphics[height=3cm]{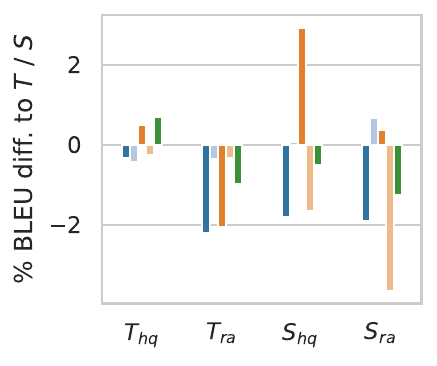}
    \includegraphics[height=3cm]{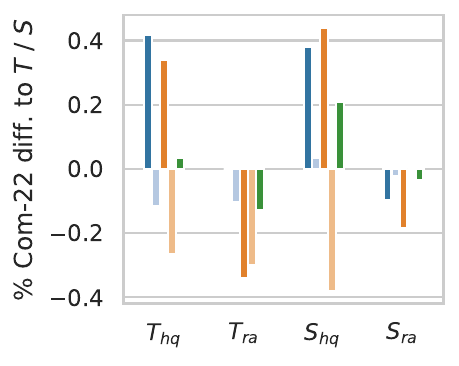}
    \includegraphics[height=3cm]{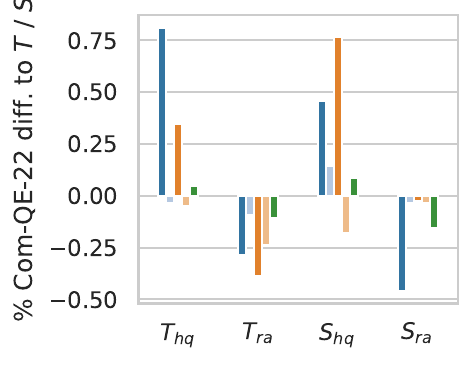}
    \includegraphics[height=3cm]{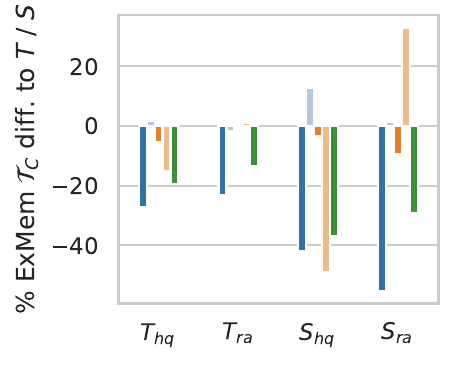}

    \includegraphics[height=3cm]{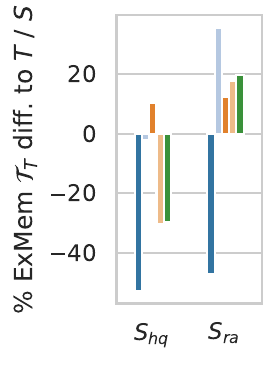}
    \includegraphics[height=3cm]{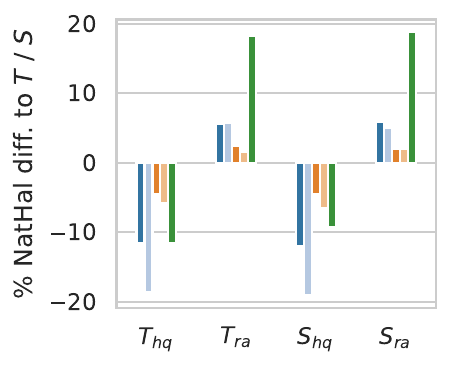}
    \includegraphics[height=3cm]{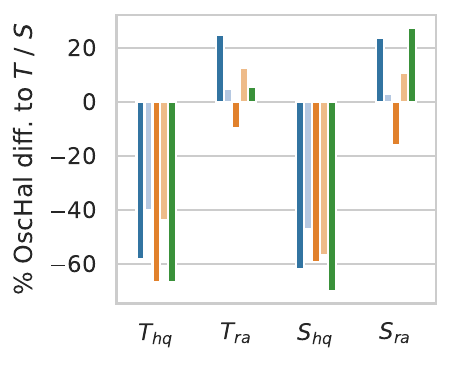}
    \includegraphics[height=3cm]{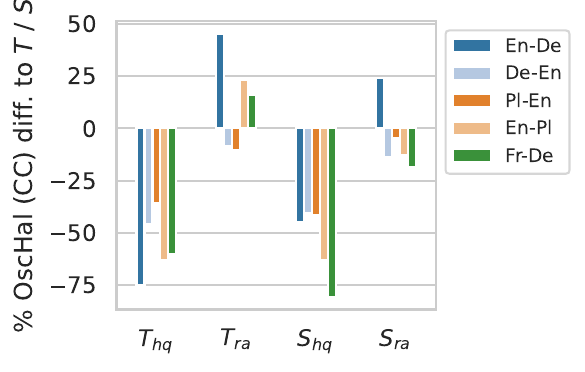}
    \end{center}
    \captionof{figure}{Performance changes observed for the different language pairs when applying Adaptive-SeqKD. Changes are computed as percentages with respect to the original teacher $\theta_T$ and $\theta_S$.}
    \label{fig:adaptivekd_perlanguage}
\end{minipage}
\end{center}

\section{Extractive memorization in commercial systems}
\label{ap:commercial}
In the main paper, we discussed ExMem for the translation systems we trained, but it should be noted that ExMem does not only exist for sytems trained for academic purposes.
To assert this, we assessed closed-source translation systems (\texttt{Google Translate}, \texttt{DeepL} and \texttt{Microsoft Bing Translator}),\footnote{As accessed on the 30th of March, 2025. For \texttt{Microsoft Bing Translator}, the `formal tone' was used for translation.} using 70 Europarl examples that led to ExMem for our WMT20 English-German models. ExMem is defined with respect to the training targets, and we do not know what closed-source translation systems are trained on. Still, because of how widespread Europarl is in translation corpora, it may have been a part of the training material for closed-source systems, too. For \texttt{DeepL}, the system outputs the target translation without having processed the full source sequence for 25 of those 70 examples. For \texttt{Google Translate} and \texttt{Microsoft Bing Translator}, that applies to 16 and 20 examples, respectively.

Four examples are provided below. In each source text below, if we omit the italicised part in square brackets, the translations remain the same. Consider the following two examples from \texttt{DeepL}:

{
\setlength{\Extopsep}{0.5em}
\setlength{\Exlabelsep}{0.75em}
\setlength{\Exlabelwidth}{.75em}
\setlength{\SubExleftmargin}{1.2em}

\ex. \label{ex1deepl}
\a.[$s$] I have received seven motions for resolution tabled in accordance with Rule 103(2) \textit{[of the Rules of Procedure]}
\b.[$t$] Ich habe sieben Entschließungsanträge erhalten, die gemäß Artikel 103 Absatz 2 der Geschäftsordnung eingereicht wurden

\ex. \label{ex2deepl}
\a.[$s$] I have therefore abstained \textit{[from the vote]}
\b.[$t$] Ich habe mich daher der Stimme enthalten

\noindent Consider the following example from \texttt{Microsoft Bing Translator}:
\ex. \label{ex1msft}
\a.[$s$] Mr President, the opposite is \textit{[the case]}
\b.[$t$] Herr Präsident, das Gegenteil ist der Fall

\noindent Consider the following example from \texttt{Google Translate}:
\ex. \label{ex1gt}
\a.[$s$] Mr President, Madam Commissioner, ladies \textit{[and gentlemen]}
\b.[$t$] Herr Präsident, Frau Kommissarin, meine Damen und Herren

}
These are examples copied directly from Europarl, but the underlying robustness issue also pops up when using non-exact variants -- e.g. consider the following variant of Example~\ref{ex1gt}: “A cleaner, a teacher and a commissioner said: ladies and”. \texttt{Google Translate} still translates this including ``Meine Damen und Herren'' in its translation. It is hard to determine in which other domains ExMem occurs for these models, because we do not have a lot of insight into their training data, but that it is not something that only applies to our models, is for certain.

It is vital that we learn to understand what causes this type of memorization and how to avoid it. We identified knowledge distillation as contributing to exacerbating these types of memorization. We hope our insights can help both scientists and practitioners tackle such problems across different production systems as well, especially since deployed systems tend to be distilled (student) models.

\section{Subgroup analyses}
\label{ap:subgroups}

Before elaborating on the results for the subgroup analysis, we more elaborately explain how we approximated the CM scores.

\subsection{Composing the counterfactual memorization subgroups}
\label{ap:cm}
Assuming input $x$ and target $y$, and a model with the parameters $\theta^{\text{tr}}$ trained on all training data, and $\theta^{\text{tst}}$ trained on all examples except $(x,y)$, CM can be computed as follows \citep{feldman2020does,feldman2020neural}:
\begin{equation*}
\vspace{-0.1cm}
    \texttt{CM}(x, y) {=} \underbrace{p_{\theta^{\text{tr}}}(y|x)}_{\texttt{IN}} - \underbrace{p_{\theta^{\text{tst}}}(y|x)}_{\texttt{OUT}}
\end{equation*}
Leaving out individual datapoints, as this equation suggests, is computationally too expensive given the vast sizes of our WMT20 datasets.
We, therefore, approximate CM scores for all datapoints for \textsc{En}-\textsc{De}, and for 10\% of the datapoints for the remaining language pairs. Following \citet[][p.3]{dankers2023memorisation}, who previously computed CM scores for NMT examples, we replace the probability of the full target with the geometric mean of the target token probabilities.
This is more robust to length differences than the full target probability.

For \textsc{En}-\textsc{De}, we approximate CM by training 10 teacher models on a randomly sampled 50\% of the training corpus, while evaluating it on the remaining 50\%, such that for each datapoint, the `\texttt{IN}' and `\texttt{OUT}' quantities in this equation are both estimated using five models.

For the remaining language pairs, we use the original $\theta_T$ model to estimate the `\texttt{IN}' quantity, and train another model according to the same training procedure on 90\% of the training data, to estimate the `\texttt{OUT}' quantity for 10\% of the remaining training data. This is a rather coarse estimation, but should suffice to determine generic relations between CM and our evaluation metrics of interest.

Using the CM scores, we create six subgroups of interest. The last four were also contained in the main paper, and here we add the first two, that separate examples with a low CM score into two groups:
\begin{itemize}[noitemsep, topsep=0pt]
    \item \texttt{IN} and \texttt{OUT} performances $\leq0.2$, marked `L(ow)-L(ow)' in the figures;
    \item \texttt{IN} and \texttt{OUT} performances $\geq0.8$, marked `H(igh)-H(igh)' in the figures;
    \item $\{[0, 0.2), [0.2, 0.3), [0.3, 0.4), [0.4, 1.0]\}$. 
\end{itemize}

\subsection{Results}

\paragraph{Random subgroup} Our baseline subgroup consists of 50k examples randomly sampled from each of the WMT20 corpora.
Replication metrics (chrF in Figure~\ref{fig:random_subgroup}) follow the overall patterns of $\theta_T$>$\theta_S$>$\theta_B$, consolidating that SeqKD dampens memorization in $\theta_S$ compared to $\theta_T$ but amplifies it compared to $\theta_B$.
The quality and diversity metrics (Comet-QE-22 and MSTTR in Figure~\ref{fig:random_subgroup}) emphasize that, compared to the corpus (column $\mathcal{T}_C$), the models generate higher-quality translations, but with lower textual diversity.

\vspace{0.3cm}

\noindent\begin{minipage}{\textwidth}
    \centering
    \includegraphics[width=0.85\textwidth]{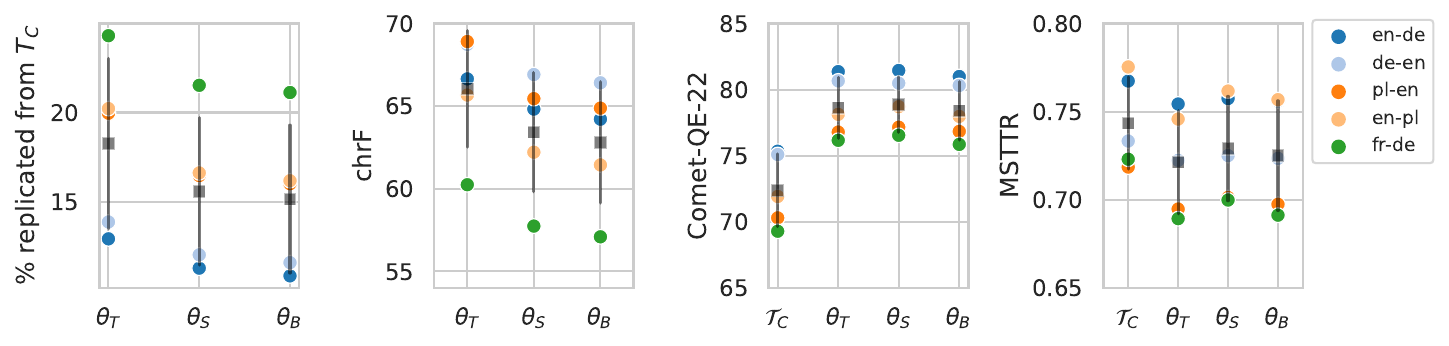}
    \captionof{figure}{Evaluation metrics applied to the random subgroup, for all five language pairs. The square marker indicates the mean and standard deviation.}
    \label{fig:random_subgroup}
\end{minipage} 
\vspace{0.3cm}

\paragraph{Quality-based subgroups} How does KD affect examples from the WMT20 corpus with a certain quality? To categorize WMT20 examples based on quality, we applied the Comet-QE-22 metric using the corpus's targets as translations (because the reference is now the translation, we apply Comet's `reference-free' method).
We examine five subgroups: $\{[0, 0.2), [0.2, 0.4), [0.4, 0.6), [0.6,0.8), [0.8, 1.0]\}$.

In Figure~\ref{fig:comet_subgroups}, we first notice that for the low-quality subgroups, fewer translations are replicated from $\mathcal{T}_C$ (all models) and from $\mathcal{T}_T$ (for $\theta_S$, see Figure~\ref{fig:replication_tt_cqe}).
Secondly, for groups with a quality below 0.6, both $\theta_S$ and $\theta_B$ generated better translations than $\theta_T$, as per Comet-QE-22 applied to the model-generated translations. 
Textual diversity metrics here mostly follow the same trend as Comet-QE-22, with $\theta_S$ showing the highest diversity.

Figure~\ref{fig:subgroups_rel_diffs} visualizes the relative differences of students/baselines to the teacher more explicitly, as a percentual increase.
For Comet-QE-22, if the student's bar exceeds the baseline's bar, this signifies that the improvement over the teacher is partially attributable to the capacity gap between large and base models, and partially to the SeqKD process.
This improvement over the teacher holds for the lower quality groups (0-0.2, 0.2-0.4, 0.4-0.6), but not for the higher quality groups.
Example~\ref{ex4} from the introduction already illustrated a sample from the lowest quality subgroup: in the corpus, the target was likely misaligned. $\theta_T$'s translation is slightly better but still wrong, and $\theta_S$'s translation is slightly better than $\theta_B$'s.
Compared to $\theta_B$, $\theta_S$ benefits from being presented with the teacher's corpus, which is a \textbf{denoised} version of WMT20, since $\theta_T$ replicates the least for the lowest-quality (noisiest) examples.
$\theta_S$, in turn, replicates less from $\mathcal{T}_T$ for these lowest-quality subgroups, too, which allows for \textbf{amplified denoising} compared to $\theta_T$.

\vspace{0.3cm}
\noindent\begin{minipage}{\textwidth}
    \includegraphics[width=\textwidth]{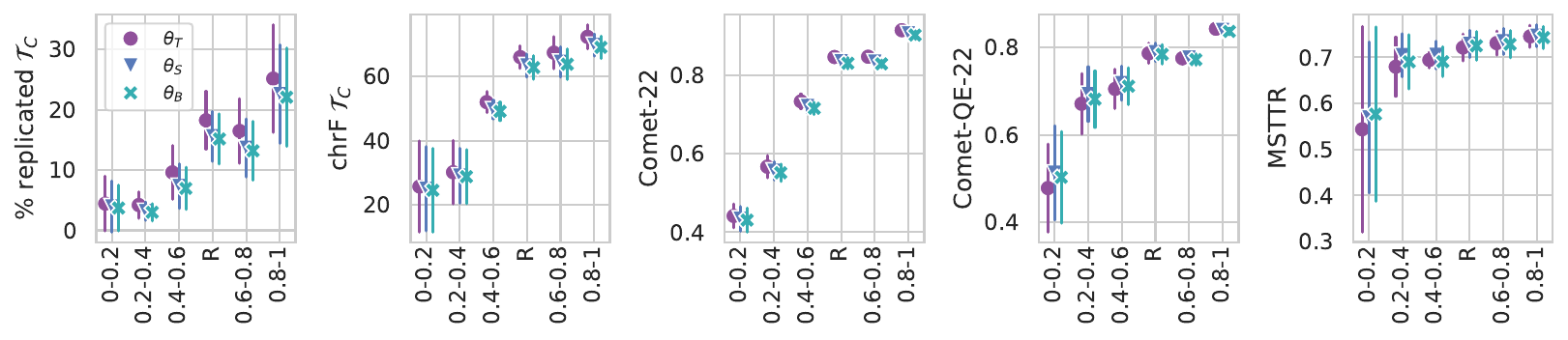}
    \captionof{figure}{Evaluation metrics applied to the quality subgroups, aggregated over language pairs with error bars indicating standard deviation over language pairs.}
    \label{fig:comet_subgroups}
\end{minipage}
\vspace{0.3cm}

\paragraph{Counterfactual memorization subgroups}
If we now focus on the high CM subgroups (introduced in the previous subsection) in Figure~\ref{fig:cm_subgroups}, we firstly observe that in terms of memorization metrics, the teacher shows increased replication and chrF. This follows from the definition of CM as the \texttt{IN} metric is expected to correlate highly with replication: the higher the target probability, the more likely it is that the teacher can replicate the target when generating translations. We note, however, that the student and baseline replicate examples with high CM less than examples with lower CM. This is likely due to their lower capacity: by definition, CM highlights examples for which a model assigns a low probability to the target when that example is not in the training set, thus requiring more capacity/parameters from the original $\theta_T$ to learn that target.
The reduced replication also leads to lower Comet-22 scores for these examples, since that metric partially relies on the corpus's target.

When looking at the Comet-QE-22 scores, we observe that the higher CM groups typically also have lower Comet-QE-22 scores, although the quality scores are still above the lowest quality subgroup discussed in the previous paragraph.
The baseline and student models outperform $\theta_T$ here, likely because they struggle to replicate the somewhat noisy targets as well as the teacher did. The student shows some amplified denoising compared to $\theta_B$, but that does apply to CM groups across the board, and does not seem specific to individual CM subgroups.

All three models show more textual diversity for subgroups with higher CM scores.

Finally, if we inspect the two groups with very low and very high \texttt{IN} and \texttt{OUT} scores, the models---as expected based on the CM definition---do not replicate the `low' group, and have very high replication rates for the `high' group. For the remaining metrics, the `low' group underscores the findings we previously reported for the low-quality subgroups: the student model shows amplified denoising, as reflected by the Comet-QE-22 metric, and also shows more textual diversity than the other two models.
Examples that score badly in terms of both \texttt{IN} and \texttt{OUT} performance are typically low-quality, misaligned examples \citep{dankers2023memorisation}; if the teacher does not replicate those low-quality targets, but partially denoises by translating them more accurately than the corpus did, the student can further improve upon that.

\vspace{0.3cm}

\noindent\begin{minipage}{\textwidth}
    \includegraphics[width=\textwidth]{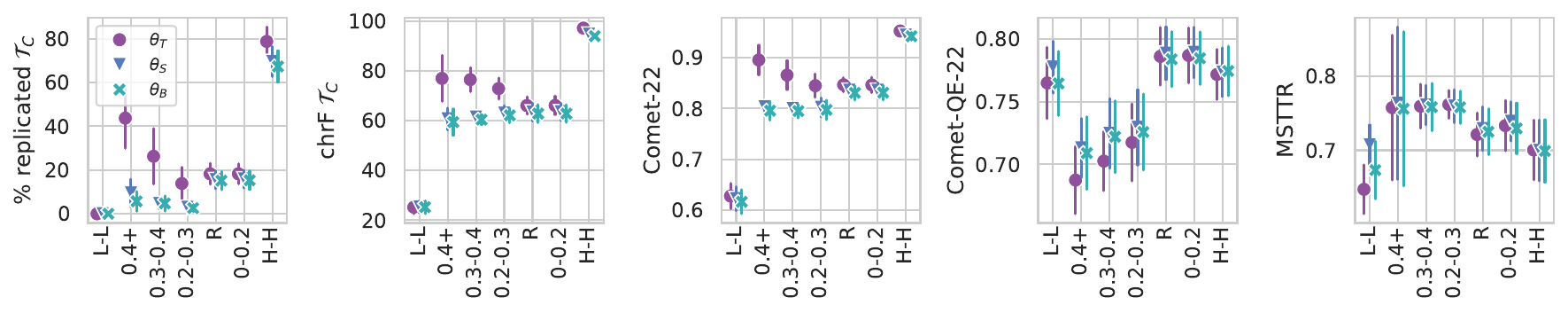}
    \captionof{figure}{Evaluation metrics applied to the CM subgroups, aggregated over language pairs with error bars indicating standard deviation over language pairs.}
    \label{fig:cm_subgroups}
\end{minipage}
\vspace{0.05cm}

\paragraph{Confidence subgroups}
Finally, we turn to the confidence-based groups, created using the lowest 10k and highest 10k examples based on the mean token-level log-probability of the translations the teacher generated. Figure~\ref{fig:confidence_subgroups} demonstrates that the low confidence examples stand out for all metrics: none of the models replicate their targets, all three models generate translations with a relatively low Comet-QE-22 score (akin to the lowest-quality subgroups discussed above), and all three models show an exceptionally low MSTTR. If we inspect the examples, they contain many misaligned examples from the corpus for which the teacher also does not generate an adequate translation (e.g., Example~\ref{ex6}), and a lot of cases where the teacher hallucinates in its translation (e.g., Example~\ref{ex7}). The student's improvement here thus does not stem from learning from the superior targets $\theta_T$ generated but more likely from deviating from $\theta_T$. Yet, note that even if $\theta_S$ does not necessarily copy these hallucinations from its teacher, we did find that the students inspected in the main paper showed amplified hallucination capabilities, so low-quality teacher targets like this could have other unanticipated effects downstream.

{
\setlength{\Extopsep}{0.5em}
\setlength{\Exlabelsep}{1.2em}
\setlength{\Exlabelwidth}{.75em}
\setlength{\SubExleftmargin}{1.5em}

\ex. \label{ex6}
\a.[$s_C$] Evergreen Terrace - Almost Home 22.
\b.[$t_C$] Zum Ernst-Gettke-Haus, Hausnummer 68, siehe unten.
\c.[$t_T$] Die alte Burganlage – heute noch 22 Hektar.
\d.[$t_S$] Die Evergreen Terrace ist fast die Heimat 22.
\e.[$t_B$] In: Evergreen Terrace – Fast Home 22.

\ex. \label{ex7}
\a.[$s_C$] Because the race is restricted to Canadian-bred horses, it is not eligible for grading, despite being one of Canada's most prestigious races Northern Dancer Turf Stakes, (\dots)
\b.[$t_C$] Kammerherr, Land- und Obergerichtsrat Magnus Graf von Moltke, Ständedeputierter der Stadt Schleswig, Präsident der konstituierenden Ständeversammlung des Jahres 1836, (\dots)
\c.[$t_T$] Stakes, Stakes, Stakes, Stakes, Stakes, Stakes, Stakes, Stakes, Stakes, Stakes, Stakes, Stakes, Stakes, Stakes, Stakes, Stakes, Stakes, Stakes, Stakes, Stakes, Stakes, (\dots)
\d.[$t_S$] Depots of Canada'hintergrund für kanadische Pferde, ist es nicht berechtigt für das Einstufung, obwohl es eines der renommiertesten Rennen Northern Dancer Turf Stakes, (\dots)
\e.[$t_B$] Stakes, Stakes, Stakes, Stakes, Stakes, Stakes, Stakes, Stakes, Stakes, Stakes, Stakes, Stakes, Stakes, Stakes, Stakes, Stakes, Stakes, Stakes, Stakes, Stakes, Stakes, (\dots)

}

\vspace{0.3cm}

\noindent\begin{minipage}{\textwidth}
    \includegraphics[width=\textwidth]{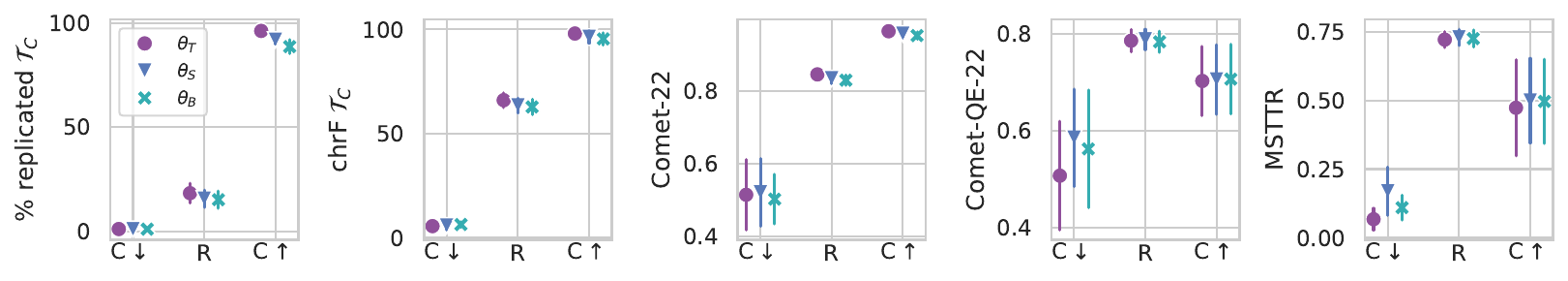}
    \captionof{figure}{Evaluation metrics applied to the confidence subgroups, aggregated over language pairs with error bars indicating standard deviation over language pairs.}
    \label{fig:confidence_subgroups}
\end{minipage}
\vspace{0.3cm}

\noindent\begin{minipage}{\textwidth}
    \centering
    \includegraphics[width=0.6\textwidth]{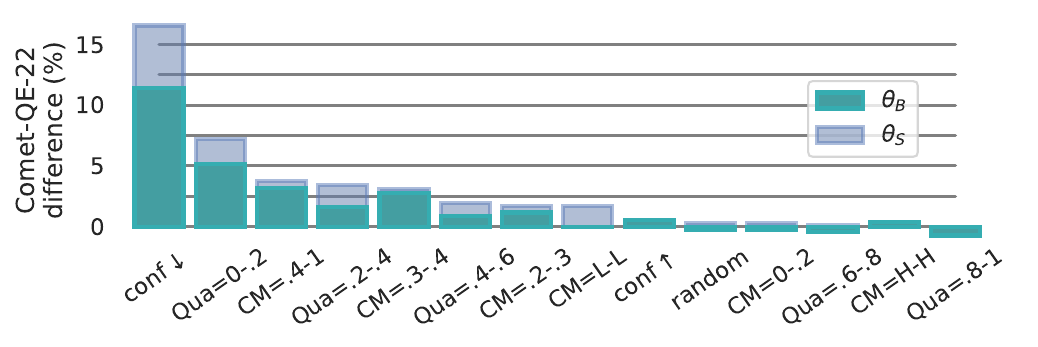}
    \captionof{figure}{Relative increases comparing students and baselines to the teacher models, for the Comet-22-QE metric.}
    \label{fig:subgroups_rel_diffs}
\end{minipage}

\end{document}